\documentclass[runningheads]{llncs}
\usepackage{graphicx}
\usepackage[ruled,linesnumbered,vlined]{algorithm2e}
\usepackage{amsmath}
\usepackage{amssymb}
\usepackage{tikz}\usetikzlibrary{arrows,fit,shapes.symbols,patterns}
\usepackage{xspace}
\usepackage{url}

\usepackage{listings}
\makeatletter
\lst@Key{countblanklines}{true}[t]{\lstKV@SetIf{#1}\lst@ifcountblanklines}
\lst@AddToHook{OnEmptyLine}{%
    \lst@ifnumberblanklines\else%
    \lst@ifcountblanklines\else%
    \advance\c@lstnumber-\@ne\relax%
    \fi%
    \fi}
\makeatother
\lstdefinelanguage{asp}{
    morekeywords={not, count, sum, min, max},
    morecomment=[l]{\%},
    breakatwhitespace=true,
    captionpos=b,
    numbers=left,
    numbersep=5pt,
    numberblanklines=false,
    countblanklines=false,
    commentstyle=\color{gray},
    frame=bt, framexbottommargin=5pt, framextopmargin=5pt,
    aboveskip=5pt, belowskip=5pt,
    abovecaptionskip=10pt
}
\lstnewenvironment{asp}{
	\lstset{
		language=asp,
		showstringspaces=false,
        keepspaces=true,
		formfeed=\newpage,
		tabsize=4,
		commentstyle=\color{blue},
		numbers=none,
		breaklines=true,
		literate={~} {$\sim$}{1},
    frame=none,
	}
}{}
\def\la{\ensuremath{\textnormal{ :-- }}}
\newcommand{\tuple}[1]{\ensuremath{\left \langle #1 \right \rangle }}
\newcommand{\magicRules}[1]{\ensuremath{R^{\it mgc}}\xspace}
\newcommand{\modifiedRules}[1]{\ensuremath{R^{\it mod}}\xspace}

\newcommand{\naf}{\ensuremath{\mathtt{not}}\xspace}

\usepackage{multirow}

\newcommand{\NC}{\ensuremath{N_C}\xspace}
\newcommand{\NR}{\ensuremath{N_R}\xspace}
\newcommand{\NI}{\ensuremath{N_I}\xspace}

\newcommand{\Amc}{\ensuremath{\mathcal{A}}\xspace}

\newcommand{\Kmc}{\ensuremath{\mathcal{K}}\xspace}

\newcommand{\Tmc}{\ensuremath{\mathcal{T}}\xspace}

\newcommand{\dlliter}{$\mathsf{DL}$-$\mathsf{Lite_{R}}$\xspace}

\newcommand{\elh}{\ensuremath{\mathcal{E\!LH}}\xspace}

\newcommand{\hshiq}{Horn-$\mathcal{SHIQ}$\xspace}

\newcommand{\T}{\mathcal{T}}
\newcommand{\A}{\mathcal{A}}

\newcommand{\dlisa}{\sqsubseteq}

\newcommand{\nop}[1]{}

\begin{document}
\title{Large-scale Ontological Reasoning via Datalog}
\author{Mario Alviano\orcidID{0000-0002-2052-2063} \and
Marco Manna\orcidID{0000-0003-3323-9328}}
\authorrunning{M. Alviano \and M. Manna}
\institute{University of Calabria, 87036 Rende, Italy\\
\email{\{alviano,manna\}@mat.unical.it}}

\maketitle              % typeset the header of the contribution

\begin{abstract}
Reasoning over OWL 2 is a very expensive task in general, and therefore the W3C identified tractable profiles exhibiting good computational properties.
Ontological reasoning for many fragments of OWL 2 can be reduced to the evaluation of Datalog queries.
This paper surveys some of these compilations, and in particular the one addressing queries over Horn-$\mathcal{SHIQ}$ knowledge bases and its implementation in DLV2 enanched by a new version of the Magic Sets algorithm.
\keywords{Datalog  \and Ontology Reasoning \and Query Answering.}
\end{abstract}

\section{Introduction}
% (2 pagine incluso abstract):  Marco et al.

Datalog is a rule-based language originally designed in the context of {\em deductive databases}, a field that has benefited from the cross-fertilization between logic programming and database theory \cite{DBLP:books/aw/AbiteboulHV95,DBLP:journals/tkde/CeriGT89,DBLP:books/sp/CeriGT90,DBLP:books/sp/Lloyd87}.
The language is nowadays successfully applied in several contexts, spanning from Boolean optimization and constraint satisfaction problems~\cite{DBLP:journals/jcss/BodirskyD13} to ontological design and reasoning in the Semantic Web~\cite{DBLP:journals/ws/CaliGL12,DBLP:conf/ijcai/GrauKKZ15}.
Specifically, standard reasoning tasks in the Semantic Web are often reduced to query evaluation over (deductive) databases, so to satisfy the fundamental prerequisite of efficient large-scale reasoning.

This paper focuses on the use of Datalog in the context of Semantic Web, in particular on its application to {\em ontology-based query answering}, for short OBQA~\cite{DBLP:conf/dlog/CaliGL09,DBLP:conf/amw/Ortiz13}.
In OBQA, a Boolean query $q$ has to be evaluated against a {\em logical theory} (a.k.a. {\em knowledge base}, or KB) consisting of an extensional {\em database} (a.k.a. ABox) $D$ paired with an {\em ontology} (a.k.a. TBox) $\Sigma$.
The problem is usually stated as $D \cup \Sigma \models q$, and is equivalent to checking whether $q$ is satisfied by all models of $D \cup \Sigma$ according to the classical {\em open-world assumption} (OWA) of first-order logic~\cite{DBLP:books/aw/AbiteboulHV95}.
Several fields of Computer Science have shown interest in OBQA, from
Artificial Intelligence~\cite{DBLP:journals/ai/BagetLMS11,%
	DBLP:journals/ai/CalvaneseGLLR13,%
	DBLP:journals/ai/GottlobKKPSZ14} %%AI/KR
to Database Theory~\cite{DBLP:journals/tods/BienvenuCLW14,%
	Bourhis:2016:GDT:3014437.2976736,DBLP:journals/tods/GottlobOP14} %%DB TODS
and Logic~\cite{%
%PerezUrbina2010186,%
DBLP:journals/corr/BaranyGO13,%
	DBLP:conf/icalp/GottlobPT13ICALP,perez2010tractable}. %%LOGIC
From these fields, two families of formal knowledge representation languages to specify $\Sigma$ emerged, namely Description Logics (DLs)~\cite{Baader:2003:DLH:885746} and Datalog$^\pm$~\cite{DBLP:conf/dlog/CaliGL09}.
For both of them, OBQA is undecidable in general \cite{DBLP:journals/jair/CaliGK13,DBLP:journals/jcss/JohnsonK84,DBLP:conf/icdt/Rosati07},
and therefore syntactic decidable fragments have been singled out with the aim of offering a good balance between computational complexity and expressiveness.
The same idea lead to the definition of \emph{OWL 2 Web Ontology Language Profiles}.

In fact, reasoning over OWL 2 is generally a very expensive task:
fact entailment (i.e., checking whether an individual is an instance of a concept) is already 2\textsc{NExpTime}-hard, while decidability of conjunctive query answering is still an open problem.
To balance expressiveness and scalability, the W3C identified three tractable profiles, namely OWL 2 EL, OWL 2 QL, and OWL 2 RL, exhibiting good computational properties:
the evaluation of conjunctive queries over KBs falling in these fragments is in \textsc{PTime} in data complexity (that is, when query and TBox are fixed) and in \textsc{PSpace} in combined complexity (that is, in the general case)~\cite{DBLP:journals/jair/StefanoniMKR14}.
Horn-$\mathcal{SHIQ}$ is another fragment of OWL 2 exhibiting good computational properties and high expressivity:
conjunctive queries are evaluated in \textsc{PTime} in data complexity, and in \textsc{ExpTime} in combined complexity;
it generalizes both OWL 2 QL and OWL 2 RL, and captures all OWL 2 EL constructs but role chain~\cite{DBLP:conf/ic3k/Leone18}.

From a theoretical viewpoint much has been done:
OBQA has been addressed in many ontological settings by reductions to the evaluation of Datalog queries \cite{DBLP:conf/kr/CarralDK18,DBLP:conf/aaai/EiterOSTX12,DBLP:conf/ijcai/KontchakovLTWZ11,DBLP:conf/dlog/StefanoniMH12,DBLP:conf/ijcai/XiaoCKLPRZ18}.
Some of these rewriting are briefly mentioned in Section~\ref{sec:compilations}.
From a pragmatic viewpoint, reasoning services have been developed, among them \textsc{mastro}~\cite{DBLP:journals/semweb/CalvaneseGLLPRRRS11},
\mbox{\textsc{ontop}}~\cite{DBLP:journals/semweb/CalvaneseCKKLRR17},
and \textsc{rdfox}~\cite{DBLP:conf/semweb/NenovPMHWB15}.
Concerning \hshiq, Eiter et al.~\cite{DBLP:conf/aaai/EiterOSTX12} showed that OBQA can be addressed by means of Datalog queries:
in a nutshell, given a \hshiq TBox paired with a SPARQL query \cite{SPARQL_Entailment_Regimes}, it is possible to construct an equivalent Datalog query independently from the ABox.
This idea has been also implemented in a specific branch of DLV2~\cite{DBLP:conf/lpnmr/AlvianoCDFLPRVZ17}, with promising results.
The rewriting implemented by DLV2 is further processed by a new version of the \emph{magic sets algorithm} \cite{DBLP:journals/tplp/AlvianoLVZ19}, which inhibits the creation of new recursive dependencies and  partially unroll the magic sets rewriting if binding information are lost.
The inhibition of new recursive definitions is important because magic sets were originally introduced for Datalog programs \cite{DBLP:conf/pods/BancilhonMSU86}, and their extension to programs with stratified negation was nontrivial;
indeed, the perfect model semantics \cite{DBLP:journals/jar/Przymusinski89} is not applicable to the rewritten program if recursive negation is introduced by magic sets, and several semantics were considered in the literature to overcome this limitation \cite{DBLP:journals/jlp/BalbinPRM91,DBLP:conf/pods/Behrend03,DBLP:journals/tcs/KempSS95,DBLP:conf/fgcs/KerisitP88,DBLP:journals/jacm/Ross94}.
By inhibiting the creation of new recursive dependencies, the semantic issue disappears.
The technique is briefly explained in Section~\ref{sec:magic}.

\section{Background}\label{sec:background}

\subsection{Basics}
Fix three pairwise disjoint discrete sets $\mathsf{C}$, $\mathsf{P}$ and $\mathsf{V}$, respectively of {\em constants}, {\em predicate symbols} and {\em variables}. Constants and variables all together form what we call {\em terms}. Each predicate symbol $p$ has an arity, consisting of a non-negative integer $\mathit{arity}(p)$. An {\em atom} is an expression $\alpha$ of the form $p({\bf t})$, where $p$ is a predicate symbol, $\mathbf{t} = t_1,...,t_m$ is a sequence of terms possibly with repetitions, $m$ is the arity of $p$, $\alpha[i] = t_i$  for each $i \in [1..m]$, and the set $\{ t_1,...,t_m\}$ is denoted by $\mathit{dom}(\alpha)$. By definition, $\mathit{arity}(\alpha) = \mathit{arity}(p)$. An instance $I$ is any set of atoms over constants.

\subsection{Description Logics and OWL}
Description Logics (DLs) are a family of formal knowledge representation languages that model concepts, roles, individuals, and their relationships.
Let \NC (\emph{concepts}), \NR (\emph{roles}) and \NI (\emph{individuals}) be mutually disjoint discrete sets. Hereinafter, we assume that
$\NC \cup \NR \subset \mathsf{P}$ and that $\NI \subset \mathsf{C}$. Accordingly, concepts are basically unary predicates whereas roles are binary predicates. Moreover, in this context, concepts and roles are  denoted by uppercase letters.
A DL \emph{knowledge base} (KB) in normal form is any pair $\Kmc=(\Amc,\Tmc)$ where:
\begin{itemize}
	\item[$(i)$] \Amc, the ABox (assertional box), is a finite set of \emph{assertions} (i.e., atoms) of the form
	$A(a)$ or $R(a,b)$, with \mbox{$a,b\in\NI$}, $A\in\NC$, and $R\in\NR$. Roughly, an ABox can be transparently seen as a database (i.e., a finite instance).
	
	\item[$(ii)$] \Tmc, the TBox (terminological box), is a finite set of {\em concept inclusions} (CIs)
	together with a finite set of \emph{role inclusions} (RIs). Table~\ref{tab:CI} and Table~\ref{tab:RI} report only those inclusions that are at the basis of the OWL 2 Web Ontology Language Profiles introduced below. Accordingly, we consider the following classes of Description Logics:
	$\mathcal{EL}$++~\cite{DBLP:conf/ijcai/BaaderBL05},
	$\mathsf{Horn}$-$\mathcal{SHIQ}$~\cite{hustadtdata},
	\elh~\cite{de2004polynomial},
	\mbox{\dlliter}~\cite{rosati2010improving}, and
	DLP~\cite{grosof2003description}.	
	The semantics of concept (resp., role) inclusions is given in Table~\ref{tab:CI} (resp., \ref{tab:RI})
	in terms of first-order expressions~\cite{Baader:2003:DLH:885746}.
	\end{itemize}

The OWL 2 Web Ontology Language, informally OWL 2, is an ontology language for the Semantic Web with formally defined meaning. OWL 2 ontologies are stored as Semantic Web documents and provide classes, properties, individuals, and data values.
The most expressive OWL 2 profile is called \mbox{OWL 2 DL.}

\renewcommand{\arraystretch}{1.3}
% Horn-SHIQ: http://www.aaai.org/Papers/AAAI/2007/AAAI07-071.pdf
\begin{table}[b!]
    \caption{Concept inclusions, where $A,B,B_1,B_2 \in \NC$ and $R \in \NR$. In the last column, all occurrences of the variables $x$, $y$, $y_1$ and $y_2$ are intended to be universally quantified. }\label{tab:CI}\normalsize
	\begin{center}\scriptsize
		\begin{tabular}{c|c|c|c|c||c|c}
			\hline
			%	$\mathcal{EL}$++
			%    %~\cite{DBLP:conf/ijcai/BaaderBL05}
			%    & $\mathsf{Horn}$-$\mathcal{SHIQ}$
			%    %~\cite{hustadtdata}
			%    & \elh
			%    %~\cite{de2004polynomial}
			%    & \dlliter
			%    %~\cite{rosati2010improving}
			%    & DLP
			%    %~\cite{grosof2003description}
			%    & concept & Equivalent \\
			%	(OWL2EL) & ($\supset$ OWL2EL) & ($\subset$ OWL2EL) & (OWL2QL) &   ($\approx$OWL2RL) & inclusions & Datalog$^\pm$ rule \\
			\multirow{2}{*}{~$\mathcal{EL}$++~}
			& \multirow{2}{*}{~\hshiq~}
			& \multirow{2}{*}{~~~\elh~~~}
			& \multirow{2}{*}{~~\dlliter~~}
			& \multirow{2}{*}{~~\,\,DLP\,\,~~}
			& concept & Equivalent \\
			&  &  &  &    & inclusions & first-order expression \\
			\hline
			$\checkmark$ & $\checkmark$	& $\checkmark$ & $\checkmark$ & $\checkmark$ &  $B \sqsubseteq A $ & $B(x) \rightarrow A(x)$\\
			\hline
			$\checkmark$ & $\checkmark$	& $\checkmark$ &  & $\checkmark$ &  \,$B_1 \sqcap B_2 \sqsubseteq A $\, & $B_1(x), B_2(x) \rightarrow A(x)$\\		
			
			\hline
			& 	\multirow{2}{*}{$\checkmark$}	& \multirow{2}{*}{} & \multirow{2}{*}{} &   \multirow{2}{*}{$\checkmark$} &  $B \sqsubseteq \forall R.A $ & 	\multirow{2}{*}{$B(x), R(x,y) \rightarrow A(y)$}\\		
			
			& 	 &  &  &  &  $\exists R^-.B \sqsubseteq A $ & \\
			\hline		
			$\checkmark$ & 	$\checkmark$ & $\checkmark$ &  & $\checkmark$ &  $\exists R.B \sqsubseteq A $ & $R(x,y), B(y) \rightarrow A(x)$\\
			\hline
			\multirow{2}{*}{$\checkmark$} & 	\multirow{2}{*}{$\checkmark$} & \multirow{2}{*}{$\checkmark$} & \multirow{2}{*}{$\checkmark$}  & \multirow{2}{*}{$\checkmark$} &  $\exists R.\top \sqsubseteq A $ & \multirow{2}{*}{$R(x,y) \rightarrow A(x)$}\\
			
			& &  &  &   &  $\mathsf{dom}(R) \sqsubseteq A $ & \\
			
			\hline	
			$\checkmark$ & $\checkmark$	&  & $\checkmark$ &  $\checkmark$ &  $\mathsf{ran}(R) \sqsubseteq A $ & $R(x,y) \rightarrow A(y)$\\
			\hline
			$\checkmark$ & $\checkmark$ & $\checkmark$ & $\checkmark$  & &  $B \sqsubseteq \exists R.A $ & $B(x) \rightarrow \exists z  R(x,z), A(x)$\\		
			\hline
			& $\checkmark$ &  &  $\checkmark$ & $\checkmark$ &  $B \sqsubseteq \neg A$ & $B(x), A(x) \rightarrow \bot$\\				
			\hline
			& \multirow{2}{*}{$\checkmark$} & \multirow{2}{*}{} & \multirow{2}{*}{} & \multirow{2}{*}{$\checkmark$} &  \multirow{2}{*}{$B \sqsubseteq \ \leqslant  1 \, R.A$} & $B(x), R(x,y_1), R(x,y_2),$ \\
			&     & & & & & \,$A(y_1), A(y_2), y_1 \neq y_2 \rightarrow \bot$\,\\				
			\hline	
		\end{tabular}
	\end{center}
\end{table}

\begin{table}[b!]
    \caption{Role inclusions, where $R,S,P \in \NR$. In the last column, all occurrences of the variables $x$, $y$ and $z$ are intended to be universally quantified.}\label{tab:RI}\normalsize	
	\begin{center}\scriptsize
		\begin{tabular}{c|c|c|c|c||c|c}
			\hline
			%	$\mathcal{EL}$++ & $\mathsf{Horn}$-$\mathcal{SHIQ}$ & \elh & \dlliter & DLP & rule & Equivalent \\
			%(OWL 2 EL) & ($\approx$ OWL 2 EL) & ($\subset$ OWL 2 EL) & (OWL 2 QL) &   ($\approx$OWL 2 RL) & inclusions & Datalog$^\pm$ rule \\
			\multirow{2}{*}{\,~$\mathcal{EL}$++~\,} & \multirow{2}{*}{~\hshiq~} & \multirow{2}{*}{~~~\elh~~~} & \multirow{2}{*}{\,~\dlliter~\,} & \multirow{2}{*}{~~~DLP~~~} & rule & Equivalent \\
			&  &  &  &    & inclusions & first-order expression \\
			\hline
			$\checkmark$ & 		$\checkmark$	& $\checkmark$ & $\checkmark$ &  $\checkmark$ &  $S \sqsubseteq R $ & $R(x,y) \rightarrow S(x,y)$\\
			\hline
			& 		$\checkmark$	&  & $\checkmark$ &  $\checkmark$ &  $S^- \sqsubseteq R $ & $S(x,y) \rightarrow R(y,x)$\\
			\hline		
			$\checkmark$ & $\checkmark$	&  &  &  $\checkmark$ &  $R^+ \sqsubseteq R $ & $R(x,y), R(y,z) \rightarrow R(x,z)$\\
			\hline		
			$\checkmark$ & 	&  &  &   &  ~$S \circ P \sqsubseteq R $~ & ~$S(x,y), P(y,z) \rightarrow R(x,z)$~\\
			\hline		
			& $\checkmark$	&  & $\checkmark$ &  $\checkmark$ &  $S \sqsubseteq \neg R $ & $S(x,y), R(x,y) \rightarrow \bot$\\
			\hline
		\end{tabular}
	\end{center}
\end{table}

Reasoning over OWL 2 DL is a very expensive task, in general.
To balance expressiveness and scalability, the World Wide Web Consortium (W3C, for short)\footnote{See \url{https://www.w3.org/}} identified also the following profiles:%
\footnote{See \url{http://www.w3.org/TR/owl2-profiles/}}
OWL 2 EL, OWL 2 QL, and OWL 2 RL, each exhibiting better computational properties.
Moreover, we point out that $\mathcal{EL}$++ is the logic underpinning OWL 2 EL, \dlliter is the logic underpinning OWL 2 QL, and DLP is the logic underpinning OWL 2 RL.
Among these three profiles, OWL 2 RL is the only one that does not
admit the usage of existential quantification in superclass expressions in the right-hand side of concept inclusions (i.e., $B \sqsubseteq \exists R.A$ in DL notation). 

\subsection{Ontology-based query answering}%(under OWA)
In this section we formally define {\em ontology-based query answering},
one of the most important ontological reasoning service
needed in the development of the Semantic Web.

A \textit{conjunctive query} is any first-order expression of the form
$q(\bar{x}) \equiv \exists \bar{y}   \  \phi(\bar{x},\bar{y}),$
where $\phi$ is a conjunction of atoms over the variables $\bar{x} \cup \bar{y}$, possibly with constants.

A {\em model} of a KB $\mathcal{K} = (\mathcal{A} ,\mathcal{T} )$ is  any instance $I \supseteq \mathcal{A}$
satisfying all the axioms of $\mathcal{T}$, written $I \models \mathcal{T}$, where  CIs and RIs, as said, can be regarded as first-order expressions.

The set of all models of $\mathcal{K}$ is denoted by $\mathsf{mods}(\mathcal{K})$.
To comply with the so-called {\em open world assumption} (OWA), note that $I$ might contain individuals that do not occur in $\mathcal{K}$.
The {\em answers} to a query $q(\bar{x})$ over an instance $I$ is the set
$q(I) = \{ \bar{a} \in \NI^{|\bar{x}|}~|~ I \models q(\bar{a})\}$ of \mbox{$|\bar{x}|$-tuples}
of individuals obtained by evaluating $q$ over $I$.
Accordingly, the  {\em certain answers} to $q$ under OWA is the set
$\mathsf{cert}(\mathcal{K},q) = \bigcap_{I \in \mathsf{mods}(D,\Sigma)} q(I).$
Finally, ontology-based query answering (OBQA) is the problem of computing $\mathsf{cert}(\mathcal{K},q)$.
	%
%Finally, the central decision problem in ontology-based query answering (OBQA), called \mbox{\QA}, can be stated as follows: {\em Given a KB $\mathcal{K}$, a conjunctive query
%	$q(\bar{x})$, and a tuple $\bar{a} \in \mathsf{N_I}^{|\bar{x}|}$, decide whether $\bar{a} \in \mathsf{cert}(\mathcal{K},q)$.}

\subsection{Datalog}\label{sec:datalog}
%TODO: definire sintassi di un programma datalog e l'operatore TP.

A Datalog program $P$ is a finite set of rules of the form
\begin{equation}\label{eq:rule}
    \alpha_0 \la \alpha_1, \ldots, \alpha_m, \naf\ \alpha_{m+1},\ldots,\naf\ \alpha_n.
\end{equation}
where $n \geq m \geq 0$, and each $\alpha_i$ is an atom;
atoms $\alpha_1, \ldots, \alpha_m$ are also called \emph{positive literals}, while $\naf\ \alpha_{m+1},\ldots,\naf\ \alpha_n$ are also called \emph{negative literals}.
A predicate $p$ occurring in $P$ is said \emph{extensional} if all rules of $P$ with $p$ in their heads are facts;
otherwise, $p$ is said \emph{intentional}.
For any \emph{expression} (atom, literal, rule, program) $E$, let $\mathit{At}(E)$ denote the set of atoms occurring in $E$.

For a rule $r$ of the form (\ref{eq:rule}), define
$H(r) := \alpha$, the \emph{head} of $r$;
$B(r) := \{\alpha_1, \ldots, \alpha_m,$ $\naf\ \alpha_{m+1},\ldots,\naf\ \alpha_n\}$, the \emph{body} of $r$;
$B^+(r) := \{\alpha_1, \ldots, \alpha_m\}$; and
$B^-(r) := \{\alpha_{m+1}, \ldots, \alpha_n\}$.
Intuitively, $B(r)$ is interpreted as a conjunction, and we will use $\alpha \la S \wedge S'$ to denote a rule $r$ with $H(r) = \alpha$ and $B(r) = S \cup S'$;
abusing of notation, we also permit $S$ and $S'$ to be literals.
If $B(r)$ is empty, the symbol $\la$ is usually omitted, and the rule is called a \emph{fact}.

A rule $r$ is \emph{safe} if every variable occurring in $r$ also occurs in $B^+(r)$.
In the following, only safe rules are considered, and programs are required to satisfy  \emph{stratification of negation}, defined next.
The \emph{dependency graph} $\mathcal{G}_P$ of a program $P$ has nodes for each predicate occurring in $P$, and an arc from $p$ to $p'$ if there is a rule $r$ of $P$ such that $p$ occurs in $H(r)$, and $p'$ occurs in $B(r)$;
the arc is marked with \naf if $p'$ occurs in $B^-(r)$.
$P$ satisfies stratification of negation if $\mathcal{G}_P$ has no cycle involving marked arcs.

\paragraph{Semantics.}
A \emph{substitution} $\sigma$ is a mapping from variables to constants;
for an expression $E$, let $E\sigma$ be the expression obtained from $E$ by replacing each variable $X$ by $\sigma(X)$.
Let $C_1,\ldots,C_n$ (for some $n \geq 1$) be the \emph{strongly connected components} (SCCs) of $\mathcal{G}_P$, sorted so that for all $1 \leq i < j \leq n$, for all $p \in C_i$ and for all $p' \in C_j$, there is no path from $p$ to $p'$ in $\mathcal{G}_P$.
Let $\mathit{heads}(P,C_i)$ denote the set of rules of $P$ whose head predicates belong to $C_i$.
The \emph{immediate logical consequence operator} of $P$ at stage $i$, denoted $T_P^i$, is defined as
\begin{align}
    T_P^i(I) := \{H(r)\sigma \mid r \in \mathit{heads}(P,C_i), B^+(r)\sigma \subseteq I, B^-(r)\sigma \cap I = \emptyset\}
\end{align}
for $i = 1..n$ and any interpretation $I$.
Let $I_0 := \emptyset$, and $I_i := T_P^i \Uparrow I_{i-1}$, for $i = 1,\ldots,n$.
The semantics of $P$ is defined as the interpretation $I_n$, in the following denoted $\mathit{TP}(P)$.

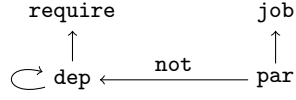
\begin{figure}[t]
    %\figrule

    \tikzstyle{node} = [text centered]
    \tikzstyle{line} = [draw, solid]
    \tikzstyle{arc} = [draw, solid, ->]
    \tikzstyle{darc} = [draw, dashed, ->]

    \centering
    \begin{tikzpicture}[scale=.9]
        \node at (0,0) [node](require) {\lstinline|require|};
        \node at (3,0) [node](job) {\lstinline|job|};
        \node at (0,-1) [node](dep) {\lstinline|dep|};
        \node at (3,-1) [node](par) {\lstinline|par|};

        \draw (dep) edge[arc] (require);
        \draw (dep) edge[loop left] (dep);
        \draw (par) edge[arc] (job);
        \draw (par) edge[arc] node[above]{\naf} (dep);
    \end{tikzpicture}
    
    \caption{
        Dependency graph of program $P_\mathit{job}$ from Example~\ref{ex:job}.
    }\label{fig:job}
    %\figrule
\end{figure}

\begin{example}\label{ex:job}
Consider a set of jobs to be executed.
Some jobs can be started only after some other jobs terminate.
Pairs of jobs that can be potentially run in parallel are identified by the following program $P_\mathit{job}$:
\begin{asp}
  dep(X,Y) :- require(X,Y).
  dep(X,Y) :- require(X,Z), dep(Z,Y).
  par(X,Y) :- job(X), job(Y), not dep(X,Y), not dep(Y,X).
\end{asp}
The dependency graph $\mathcal{G}_{P_\mathit{job}}$ is shown in Figure~\ref{fig:job}.
A possible order for the predicates of $P_\mathit{job}$ is
\lstinline|job|, \lstinline|require|, \lstinline|dep|, \lstinline|parallel|.
Given the database $D_\mathit{job}$
\begin{asp}
  job(a). job(b).          require(a,b).
  job(c). job(d). job(e).  require(c,d). require(d,e).
\end{asp}
the interpretation $\mathit{TP}(D \cup P_\mathit{job})$ extends $D$ (stage 1 and 2) with
\begin{asp}
  dep(a,b). dep(c,d). dep(d,e). $\textnormal{(stage 3, application 1)}$
  dep(c,e). $\textnormal{(stage 3, application 2)}$
  par(a,c). par(a,d). par(a,e). par(b,c). par(b,d). par(b,e).
  par(c,a). par(d,a). par(e,a). par(c,b). par(d,b). par(e,b).
\end{asp}
where the last two lines are obtained at stage 4, with one application of $T_{P_\mathit{job}}^4$.
\hfill$\blacksquare$
\end{example}

\section{Ontology Reasoning via Datalog}\label{sec:reasoning}

\subsection{Datalog Compilations}\label{sec:compilations}

For the main DL fragments described in the previous sections, OBDA can be performed via rewriting the knowledge base and the query into a Datalog program (a database and a set of Datalog rules possibly including stratified negation or negative constraints) where a special {\em output predicate} collects all the answers. 

More specifically, from a knowledge base $\mathcal{K} = (\mathcal{A} ,\mathcal{T} )$ and a conjunctive query $q(\bar{x})$, the general approach is to rewrite: $(i)$ both $\A$ and $\T$ into a database $D$; and $(ii)$ both $\T$ and $q(\bar{x})$ into a Datalog program $P$ with an output predicate $\mathsf{goal}$ of arity $|\bar{x}|$ in such a way that, $\mathsf{cert}(\mathcal{K},q)$ = $\{ \bar{a} ~|~ \mathsf{goal}(\bar{a}) \in TP(D\cup P) \}$. (Note that predicate $\mathsf{goal}$ never occurs in rule-bodies.)

We can distinguish, however, different cases. In  general, this rewriting method goes under the name of {\em combined approach}; whereas in case $D$ is independent from $\T$ (i.e., $D$ coincides with $\A$), we talk about a {\em pure approach}. Moreover, in both cases, we can distinguish two sub-cases, namely whether $P$ is an arbitrary Datalog program or it is non-recursive (which is equivalent to a union of conjunctive queries). 

For complexity and expressiveness reasons, the pure approach into non-recursive Datalog is only possible for \dlliter. 
Moreover, we also know that the combined approach into non-recursive Datalog is possible for \elh. For all the other fragments, to the best of our knowledge, what we know from the 
literature is that the rewriting methods (pure or combined) target full Datalog.

In the last decade, multiple rewriting algorithms have been proposed, even for the same DL fragment. They may differ, apart from the rewriting methods, also from the size of the rewritings, the time used to compute them, and their quality measured in terms of time and space needed during the evaluation process over classical benchmarks.

In~\cite{DBLP:conf/aaai/EiterOSTX12}, the authors provide a pure Datalog rewriting for query answering over \hshiq. In particular, in the generated program $P$, only the rules having $\mathsf{goal}$ in the head depend both on $\T$ and $q$; conversely, all the other rules of $P$ only depend on $\T$.
Program $P$ contains only unary and binary predicates, and all the rules without the $\mathsf{goal}$ predicate contain a constant number of variables. However, due to the exponential time complexity of query answering in Horn-$\mathcal{SHIQ}$, the size of $P$ is, in the worst case, exponential in the size of $\T$.
Related to this work, there is also a more recent paper \cite{DBLP:conf/aaai/CarralGK19} that consider \hshiq extended with the axiom $S \circ P \sqsubseteq R$ (see Table~\ref{tab:RI}), a.k.a. Horn-$\mathcal{SRIQ}_{\sqcap}$. However, the authors here provide a pure Datalog rewriting in case of fact entailment, namely when $q$ is a ground atom $\alpha$ of the form $A(a)$ or $R(a,b)$. In this setting, $P$ contains the rule ``$\mathsf{goal}$ {\tt :-} $\alpha${\tt .}'' plus a set of other rules that only depend on $\T$. Since $P$ contains only unary and binary predicates, and since the number of variables in rule bodies is constant, $P$ can be evaluated in polynomial time. However, due to 
the double exponential time complexity of fact entailment in Horn-$\mathcal{SRIQ}_{\sqcap}$, the size of $P$ is, in the worst case, also double exponential in the size of $\T$.

In~\cite{DBLP:conf/ijcai/LutzTW09}, the authors provide a combined Datalog rewriting for a superclass of $\elh$, called $\elh_\bot^{\mathit{dr}}$, which includes also $\mathsf{ran}(R) \sqsubseteq A$ and $B \sqsubseteq \neg A$. More precisely, the TBox and the query are rewritten into an a first-order query formula, which in turn can be translated into a non-recursive Datalog program with negation. Both the database $D$ and the program $P$ can be constructed in polynomial time. Moreover, $P$ depends only on the portion of $\T$ containing role inclusions. 

Concerning \dlliter, several rewriting algorithms have been proposed. Among these, we recall Presto~\cite{DBLP:conf/kr/RosatiA10}, QuOnto~\cite{DBLP:conf/aaai/AcciarriCGLLPR05} and Requiem~\cite{DBLP:journals/japll/Perez-UrbinaMH10}. The former, produces a pure Datalog rewriting of polynomial size. Whereas, the latter ones, produce a union of conjunctive queries (i.e., a set of Datalog rules all of which have predicate $\mathsf{goal}$ in the head) of exponential size in the worst case.

Finally, concerning DLP, one may observe that a pure Datalog rewriting (including negative constraints) can be directly obtained by taking, for each concept or role inclusion, its equivalent the first-order expression and by adding to $P$ the conjunctive query as a rule having $\mathsf{goal}$ in its head. Existing rewriting approach are Orel~\cite{DBLP:conf/dlog/KrotzschMR10}, Owl{O}nt{DB}~\cite{DBLP:conf/fhies/FaruquiM12}, RDFox~\cite{DBLP:conf/semweb/NenovPMHWB15} and DReW~\cite{DBLP:conf/csemws/XiaoEH12}.

\subsection{Magic Sets}\label{sec:magic}

\begin{algorithm}[t]
    \caption{MS($Q(\mathbf{T})$: a query atom, $P$: a program)}\label{alg:ms}
    Let $\mathbf{s}$ be such that $|\mathbf{s}| = |\mathbf{T}|$, and $\mathbf{s}_i = b$ if $\mathbf{T}_i$ is a constant, and \!$f$\! otherwise, for all $i \in [1..|\mathbf{s}|]$\;
    $P' := \{Q^\mathbf{s}(\mathbf{T}).\}$\tcp*{rewritten program: start with the magic seed}
    $S := \{\tuple{Q,\mathbf{s}}\}$\tcp*{set of produced adorned predicates}
    $D := \emptyset$\tcp*{set of processed (or done) adorned predicates}
    $G := \mathcal{G}_P \cup \{\tuple{p,m\#p} \mid p$ is a predicate occurring in $P\}$\tcp*{monitor SCCs}
    \While{$S \neq D$}{
        $\tuple{q,\mathbf{s}} := $ any element in $S \setminus D$\tcp*{select an undone adorned predicate}
        \ForEach{$r \in P$ such that $H(r) = q(\mathbf{t})$ for some list $\mathbf{t}$ of terms}{
        $P' := P' \cup \{q(\mathbf{t}) \la q^\mathbf{s}(\mathbf{t}) \wedge B(r).\}$\tcp*{restrict range of variables}
            Let $(\prec,\mathit{bnd})$ be the SIPS for $r$ with respect to $\mathbf{s}$\;
            \ForEach{$\ell \in B(r)$ such that $p(\mathbf{t'}) \in \mathit{At}(\ell)$ and $p$ is an intentional predicate of $P$}{
                $G := G \cup \{\tuple{m\#p, m\#q}\}$\;
                $B := \emptyset$\tcp*{restrict SIPS to preserve SSCs}
                \ForEach{$\ell' \in B(r)$ such that $\ell' \prec \ell$ and $p'(\mathbf{t''}) \in \mathit{At}(\ell')$}{
                    \If{$\{C \cap \mathit{At}(P) \mid C \in \mathit{SCCs}(G \cup \{\tuple{m\#p, p'}\})\} = \mathit{SCCs}(\mathcal{G}_P)$\label{alg:ms-rs:ln:restrict}}{
                        $B := B \cup \{\ell'\}$;\quad $G := G \cup \{\tuple{m\#p, p'}\}$\;
                    }
                }
                Let $\mathbf{s'}$ be such that $|\mathbf{s'}| = |\mathbf{t'}|$, and $\mathbf{s'_\mathit{i}} = b$ if $\mathbf{t'_\mathit{i}}$ is a constant or belongs to $\mathit{bnd}(\ell')$ for some $\ell' \in \{H(r)\} \cup B$ such that $\ell' \prec \ell$, and $f$ otherwise, for all $i \in [1..|\mathbf{s'}|]$\;
                $P' := P' \cup \{p^\mathbf{s'}(\mathbf{t'}) \la q^s(\mathbf{t}) \wedge B.\}$\tcp*{add magic rule}
                $S := S \cup \{\tuple{p,\mathbf{s'}}\}$\tcp*{keep track of adorned predicates}
            }
        }
        $D := D \cup \{\tuple{q,\mathbf{s}}\}$\tcp*{flag the adorned predicate as done}
    }
    \Return{$P'$}\;
\end{algorithm}

The magic sets algorithm is a top-down rewriting of the input program $P$ that restricts the range of the object variables so that only the portion of $\mathit{TP}(P)$ that is relevant to answer the query is materialized by a bottom-up evaluation of the rewritten program \cite{DBLP:conf/pods/BancilhonMSU86,DBLP:journals/jlp/BeeriR91,DBLP:journals/jlp/BalbinPRM91,DBLP:conf/pods/StuckeyS94,DBLP:journals/ai/AlvianoFGL12}.
In a nutshell, magic sets introduce rules defining additional atoms, called \emph{magic atoms}, whose intent is to identify relevant atoms to answer the input query, and these magic atoms are added in the bodies of the original rules to restrict the range of the object variables.
The procedure is reported as Algorithm~\ref{alg:ms}.
The notions of adornment, magic atom, sideway information passing strategy (SIPS), and a description of the algorithm are given next.

An adornment for a predicate $p$ of arity $k$ is any string $\mathbf{s}$ of length $k$ over the alphabet $\{b,f\}$.
The $i$-th argument of $p$ is \emph{bound} with respect to $\mathbf{s}$ if $\mathbf{s}_i = b$, and \emph{free} otherwise, for all $i \in [1..k]$.
For an atom $p(\mathbf{t})$, let $p^{\mathbf{s}}(\mathbf{t})$ be the (magic) atom $m\#p\#\mathbf{s}(\mathbf{t'})$, where $m\#p\#\mathbf{s}$ is a predicate not occurring in the input program, and $\mathbf{t'}$ contains all terms in $\mathbf{t}$ associated with bound arguments according to $\mathbf{s}$.

A SIPS for a rule $r$ with respect to an adornment $\mathbf{s}$ for $H(r)$ is a pair \mbox{$(\prec,\mathit{bnd})$}, where $\prec$ is a strict partial order over $\{H(r)\} \cup B(r)$, and $\mathit{bnd}$ maps $\ell \in \{H(r)\} \cup B(r)$ to the variables of $\ell$ that are made bound after processing $\ell$.
Moreover, a SIPS satisfies the following conditions:
\begin{itemize}
\item $H(r) \prec \ell$ for all $\ell \in B(r)$ (binding information originates from head atoms);
\item $\ell \prec \ell'$ and $\ell \neq H(r)$ implies that $\ell \in B^+(r)$ (new bindings are created only by positive literals);
\item $\mathit{bnd}(H(r))$ contains the variables of $H(r)$ associated with bound arguments according to $\mathbf{s}$;
\item $\mathit{bnd}(\ell) = \emptyset$ if $\ell$ is a negative literal.
\end{itemize}

\begin{example}\label{ex:sips}
Consider program $P_\mathit{job}$ from Example~\ref{ex:job}, and suppose we are interested in determining whether jobs \lstinline|a| and \lstinline|c| could be run in parallel.
The magic atom \lstinline|par$^{bb}$(a,b)|, that is, \lstinline|m#par#bb(a,b)|, would represent such interest.
Similarly, the interest on all jobs that could be run in parallel to job \lstinline|a| is represented by the magic atom \lstinline|par$^{bf}$(a,Y)|, that is, \lstinline|m#par#bf(a)|.
There can be several SIPS $(\prec,\mathit{bnd})$ for rule
\begin{asp}
  dep(X,Y) :- require(X,Z), dep(Z,Y).
\end{asp}
w.r.t.\ the adornment $bb$.
If body literals are processed left-to-right, and binding information is always passed when possible, then
\lstinline|dep(X,Y) ${}\prec{}$ require(X,Z) ${}\prec{}$ dep(Z,Y)|,
\lstinline|$\mathit{bnd}($dep(X,Y)$) = \{X,Y\}$|,
\lstinline|$\mathit{bnd}($require(X,Z)$) = \{Z\}$| (or $\{X,Z\}$),
and \lstinline|$\mathit{bnd}($dep(Z,Y)$)$| is irrelevant.
If instead body literals are processed in parallel, then
\lstinline|dep(X,Y) ${}\prec{}$ require(X,Z)|,
\lstinline|dep(X,Y) ${}\prec{}$ dep(Z,Y)|,
\lstinline|$\mathit{bnd}($dep(X,Y)$) = \{X,Y\}$|,
and \lstinline|$\mathit{bnd}($require(X,Z)$)$| and \lstinline|$\mathit{bnd}($dep(Z,Y)$)$| are irrelevant.
\hfill$\blacksquare$
\end{example}

Algorithm~\ref{alg:ms} starts by producing the \emph{magic seed}, obtained from the predicate and the constants in the query (lines~1--2).
After that, the algorithm processes each produced adorned predicate (lines~6--7):
each rule defining the predicate is modified so to restrict the range of the head variables to the tuples that are relevant to answer the query (lines~8--9);
such a relevance is encoded by the magic rules, which are produced for all intentional predicates in the bodies of the modified rules (lines~10--18).
Note that lines 5 and 12--16 implement a restriction of SIPS guaranteeing that no SCCs of $\mathcal{G}_P$ are merged during the application of magic sets;
there, $G \cup E$ denotes the graph obtained from the graph $G$ by adding each arc in the set $E$, and $\mathit{SCCs}(G)$ is the set of SCCs of $G$.
More in detail, a graph $G$ is initialized with the arcs of $\mathcal{G}_P$ and arcs connecting each predicate $p$ with a \emph{representative magic predicate} $m\#p$ (line~5).
After that, before creating a new magic rule, elements of $B(r)$ that would cause a change in the SCCs of $G$ are discarded (lines~13--16).
Graph $G$ is updated with new arcs involving original predicates and representative magic predicates, so that it represents a superset of the graph obtained from $\mathcal{G}_{P'}$ by merging all pairs of nodes of the form $m\#p\#\mathbf{s}$, $m\#p\#\mathbf{s'}$.

\begin{example}
Consider program $P_\mathit{job}$ from Example~\ref{ex:job}, and the query \lstinline|par(a,c)|.
The algorithm first produces the magic seed \lstinline|m#par#bb(a,b)| and the modified rule
\begin{asp}
  par(X,Y) :- m#par#bb(X,Y), job(X), job(Y), not dep(X,Y), not dep(Y,X).
\end{asp}
Relevance of \lstinline|dep(a,c)| and \lstinline|dep(c,a)| is captured by the magic rules
\begin{asp}
  m#dep#bb(X,Y) :- m#par#bb(X,Y).
  m#dep#bb(Y,X) :- m#par#bb(X,Y).
\end{asp}
Hence, new modified rules are produced:
\begin{asp}
  dep(X,Y) :- m#dep#bb(X,Y), require(X,Y).
  dep(X,Y) :- m#dep#bb(X,Y), require(X,Z), dep(Z,Y).
\end{asp}
At this point, if left-to-right SIPS are used, the magic rule
\begin{asp}
  m#dep#bb(Z,Y) :- m#dep#bb(X,Y), require(X,Z).
\end{asp}
is added and the algorithm terminates;
the rewritten program is the following:
\begin{asp}
  m#par#bb(a,c).
  m#dep#bb(X,Y) :- m#par#bb(X,Y).
  m#dep#bb(Y,X) :- m#par#bb(X,Y).
  m#dep#bb(Z,Y) :- m#dep#bb(X,Y), require(X,Z).
  
  par(X,Y) :- m#par#bb(X,Y), job(X), job(Y), not dep(X,Y), not dep(Y,X).
  dep(X,Y) :- m#dep#bb(X,Y), require(X,Y).
  dep(X,Y) :- m#dep#bb(X,Y), require(X,Z), dep(Z,Y).
\end{asp}
Note that using the parallel SIPS from Example~\ref{ex:sips} would lead to the production of the magic rule
\begin{asp}
  m#dep#fb(Y) :- m#dep#bb(X,Y).
\end{asp}
and therefore to further iterations of the algorithm to process predicate \lstinline|dep| with respect to the adornment $fb$.
\hfill$\blacksquare$
\end{example}

\begin{algorithm}[t]
    \caption{FullFree($P$: a program obtained by executing magic sets)}\label{alg:full-free}
    \ForEach{$m\#p\#f \cdots f$ occurring in $P$}{
        \ForEach{$m\#p\#\mathbf{s}$ occurring in $P$ such that $\mathbf{s} \neq f \cdots f$}{
            remove all rules of $\Pi$ having $m\#p\#\mathbf{s}$ in their bodies\;
            replace $m\#p\#\mathbf{s}(\mathbf{t})$ by $m\#p\#f \cdots f$ in all rule heads of $P$\;
        }
    }
    \Return{$P$}\;
\end{algorithm}

Depending on the processed input program, and on the adopted SIPS, some predicates may be associated with adornments containing only $f$s in the rewritten program.
Essentially, this means that all instances of these predicates in $\mathit{TP}(P)$ are relevant to answer the given query.
Hence, the range of the object variables of all rules defining such predicates cannot be actually restricted, and indeed the magic sets rewriting includes a copy of these rules with a magic atom obtained from the full-free adornment.
Possibly, the magic sets rewriting includes other copies of these rules obtained by different adornments, which may deteriorate the bottom-up evaluation of the rewritten program.
Luckily, those copies can be removed if magic rules are properly modified.
The idea is that magic rules associated with predicates for which a full-free adornment has been produced have to become definitions of the magic atom obtained from the full-free adornment.
The strategy is summarized in Algorithm~\ref{alg:full-free}, and can be efficiently implemented in two steps:
a first linear traversal of the program to identify predicates of the form $m\#p\#f \cdots f$ and to flag predicate $p$;
a second linear traversal of the program to remove and rewrite rules with predicate $m\#p\#\mathbf{s}$, for all flagged predicates $p$.

\begin{example}
Consider program $P_\mathit{job}$ from Example~\ref{ex:job}, the query \lstinline|par(a,Y)|, and parallel SIPS.
The output of the magic sets algorithm is the following (rules in the order of production):
\begin{asp}
  m#par#bf(a).
  par(X,Y) :- m#par#bf(X), job(X), job(Y), not dep(X,Y), not dep(Y,X).
  m#dep#bf(X) :- m#par#bf(X).
  m#dep#fb(X) :- m#par#bf(X).
  dep(X,Y) :- m#dep#bf(X), require(X,Y).
  dep(X,Y) :- m#dep#bf(X), require(X,Z), dep(Z,Y).
  m#dep#ff :- m#dep#bf(X).
  dep(X,Y) :- m#dep#fb(Y), require(X,Y).
  dep(X,Y) :- m#dep#fb(Y), require(X,Z), dep(Z,Y).
  m#dep#ff :- m#dep#fb(Y).
  dep(X,Y) :- m#dep#ff, require(X,Y).
  dep(X,Y) :- m#dep#ff, require(X,Z), dep(Z,Y).
  m#dep#ff :- m#dep#ff.
\end{asp}
Processing the above program with Algorithm~\ref{alg:full-free} results into the program
\begin{asp}
  m#par#bf(a).
  par(X,Y) :- m#par#bf(X), job(X), job(Y), not dep(X,Y), not dep(Y,X).
  m#dep#ff :- m#par#bf(X).
  m#dep#ff :- m#par#bf(X).
  dep(X,Y) :- m#dep#ff, require(X,Y).
  dep(X,Y) :- m#dep#ff, require(X,Z), dep(Z,Y).
  m#dep#ff :- m#dep#ff.
\end{asp}
Essentially, since everything is relevant to answer the given query using parallel SIPS, the magic sets rewriting was eventually unrolled.
\hfill$\blacksquare$
\end{example}

%\section{Related Work}

\section{Conclusion}

The semantics of many constructs of Description Logics is defined in terms of First-Order Logic expressions, and these expressions are often naturally expressible in Datalog.
A prominent example is the rewriting proposed by Eiter et al.~\cite{DBLP:conf/aaai/EiterOSTX12}, which can be applied to Horn-$\mathcal{SHIQ}$ knowledge bases.
One of the clear advantages of such compilations into Datalog is the availability of many efficient reasoners, employing several state-of-the-art techniques to optimize query answering.
As a prominent example, the magic sets rewriting drives the bottom-up evaluation of Datalog programs according to the query given in input;
recently proposed improvements to the magic sets algorithm inhibit the creation of recursive definitions and partially unroll the rewriting for predicates whose extension cannot be limited.

\bibliographystyle{splncs04}
\bibliography{bibtex}

\end{document}